\documentclass{article}
\usepackage{spconf,amsmath,graphicx,subfig,color,url}
\usepackage{enumitem,cite}
\usepackage[T1]{fontenc}
\usepackage[utf8]{inputenc}
\usepackage{tabularx}
\usepackage{colortbl}
\usepackage{hhline}
\usepackage[table]{xcolor}

\newcommand{\ie}{i.e~\!}
\newcommand{\etal}{et al.~\!}

\title{Deep-Plant: Plant Identification with convolutional neural networks}

\name{Sue Han Lee$^{\star}$ \qquad Chee Seng Chan$^{\star}$ \qquad Paul Wilkin$^{\dagger}$ \qquad Paolo Remagnino$^{\ddagger}$}
\address{$^{\star}$Centre of Image \& Signal Processing, Fac. Comp. Sci. \& Info. Tech., University of Malaya, Malaysia  \\ $^{\dagger}$Dept. Natural Capital \& Plant Health, Royal Botanic Gardens, Kew, United Kingdom \\ $^{\ddagger}$Comp. \& Info. Sys., Kingston University, United Kingdom\\
\{{\it leesuehan@siswa.um.edu.my; cs.chan@um.edu.my; p.wilkin@kew.org; p.remagnino@kingston.ac.uk\}}}

\begin{document}

\maketitle

\begin{abstract}		
This paper studies convolutional neural networks (CNN) to learn unsupervised feature representations for 44 different plant species, collected at the Royal Botanic Gardens, Kew, England. To gain intuition on the chosen features from the CNN model (opposed to a 'black box' solution), a visualisation technique based on the deconvolutional networks (DN) is utilized. It is found that venations of different order have been chosen to uniquely represent each of the plant species. Experimental results using these CNN features with different classifiers show consistency and superiority compared to the state-of-the art solutions which rely on hand-crafted features.
\end{abstract}
\section{Introduction}
\label{sec:intro}

Plants are the backbone of all life on earth providing us with food and oxygen. A good understanding of plants is essential to help in identifying new or rare plant species in order to improve the drug industry, balance the ecosystem as well as the agricultural productivity and sustainability \cite{cope2012plant}. Amongst all, botanists use variations on leaf characteristics as a comparative tool for their study on plants \cite{clarke2006venation,cope2012plant}. This is because leaf characteristics are available to be observed and examined throughout the year in deciduous, annual plants or year-round in evergreen perennials

In computer vision, despite many efforts \cite{kumar2012leafsnap,kalyoncu2014geometric,kadir2013leaf,beghin2010shape, charters2014eagle,cope2010extraction} (\ie~with sophisticated computer vision algorithms) have been conducted, plant identification is still considered a challenging and unsolved problem. This is because a plant in nature has very similar shape and colour representation as illustrated in Fig. \ref{fig:sample}. Kumar \etal {\cite{kumar2012leafsnap} proposed an automatic plant species identification system namely Leafsnap. They identified plants based on curvature-based shape features of the leaf by utilizing integral measure to compute functions of the curvature at the boundary. Then, identification is done by nearest neighbours (NN). Other solutions employed geometric \cite{hall2015evaluation}, multi-scale distance matrix, moment invariants \cite{kalyoncu2014geometric}, colour, texture \cite{kadir2013leaf, beghin2010shape} and venation features \cite{charters2014eagle,cope2010extraction} to identify a plant. Although successful, one must note that the performance of these aforementioned solutions is highly dependent on the chosen set of features which are task or dataset dependent. That is, it may suffer from the dataset bias problem \cite{torralba2011unbiased}.

\begin{figure}[t]
	\centering
	\centerline{\includegraphics[height=0.5\linewidth, width=0.9\linewidth]{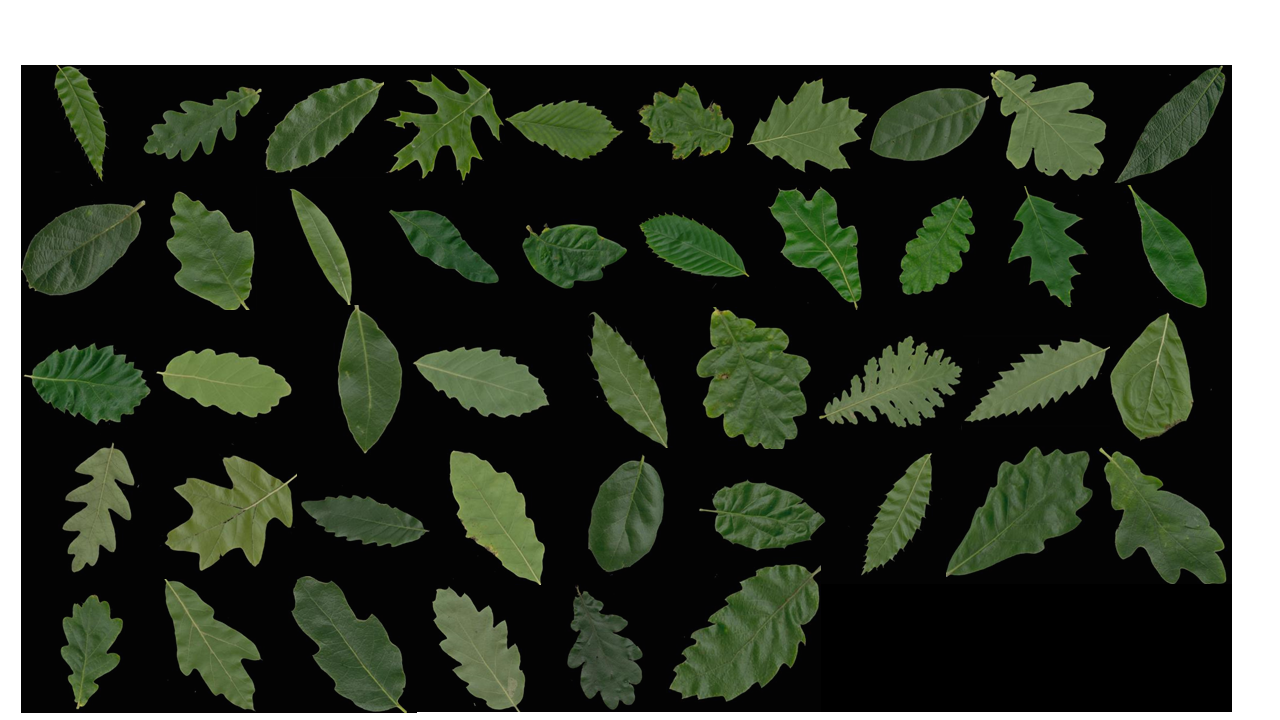}}
	\caption{Sample of the 44 plant species employed in this paper. It can be noticed that all the plant species have almost similar colour representation and shape.}
	\label{fig:sample}
\end{figure}

	\begin{figure*}[t]
	\centering
	\centerline{\includegraphics[height=0.4\linewidth, width=0.97\linewidth]{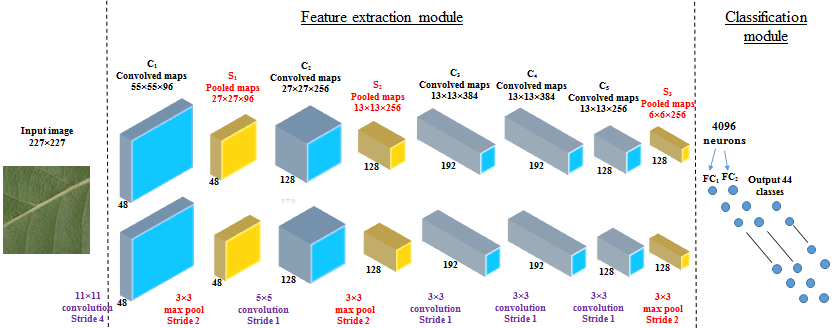}}
	\caption{Architecture of our CNN model for plant identification.}
	\label{fig:imagenet}
\end{figure*}

In this paper, we propose to employ deep learning in a bottom-up and top-down manner for plant identification. In the former, we choose to use a convolutional neural networks (CNN) model to learn the leaf features as a means to perform plant classification. In the latter, rather than using the CNN as a black box mechanism, we employ deconvolutional networks (DN) to visualize the learned features. This is in order to gain visual understanding on which features are important to identify a leaf from different classes, thus avoiding the necessity of designing hand-crafted features. Empirically, our method outperforms state-of-the-art approaches \cite{kumar2012leafsnap,yang2009linear,hall2015evaluation} using the features learned from CNN model in classifying 44 different plant species. 

This paper presents two contributions. First, we propose a CNN model to automatically learn the features representation for plant categories, replacing the need of designing hand-crafted features as to previous approaches \cite{kumar2012leafsnap, larese2014multiscale, casanova2009plant, hall2015evaluation}. Second, we identify and diagnose the feature representation learnt by the CNN model through a visualisation strategy based on the DN. This is to avoid the use of the CNN model as a black box solution, and also provide an insight to researchers on how the algorithm "see" or "perceives" a leaf. Finally, a new leaf dataset, named as MalayaKew (MK) Leaf Dataset is also collected with full annotation.

The rest of the paper is organized as follows: Section \ref{TD:TDM} reviews the concept of deep learning, in particular our CNN and DN model for plant identification. Section \ref{sec:illust} presents our findings and a comparison with conventional solutions. Finally, conclusions are drawn in Section \ref{conc}.

\section{Proposed Approach}
\label{TD:TDM}

In this section, we first explain how we employ the pre-trained CNN model to perform plant identification. Then, we detail how a DN model is utilised with our new visualisation strategy, to understand how the CNN model work in identifying different plant species. Fig. \ref{fig:deconv1} depicts the overall framework of our approach.

  \begin{figure}[t]
		\centering
		\centerline{\includegraphics[height=0.55\linewidth, width=0.98\linewidth]{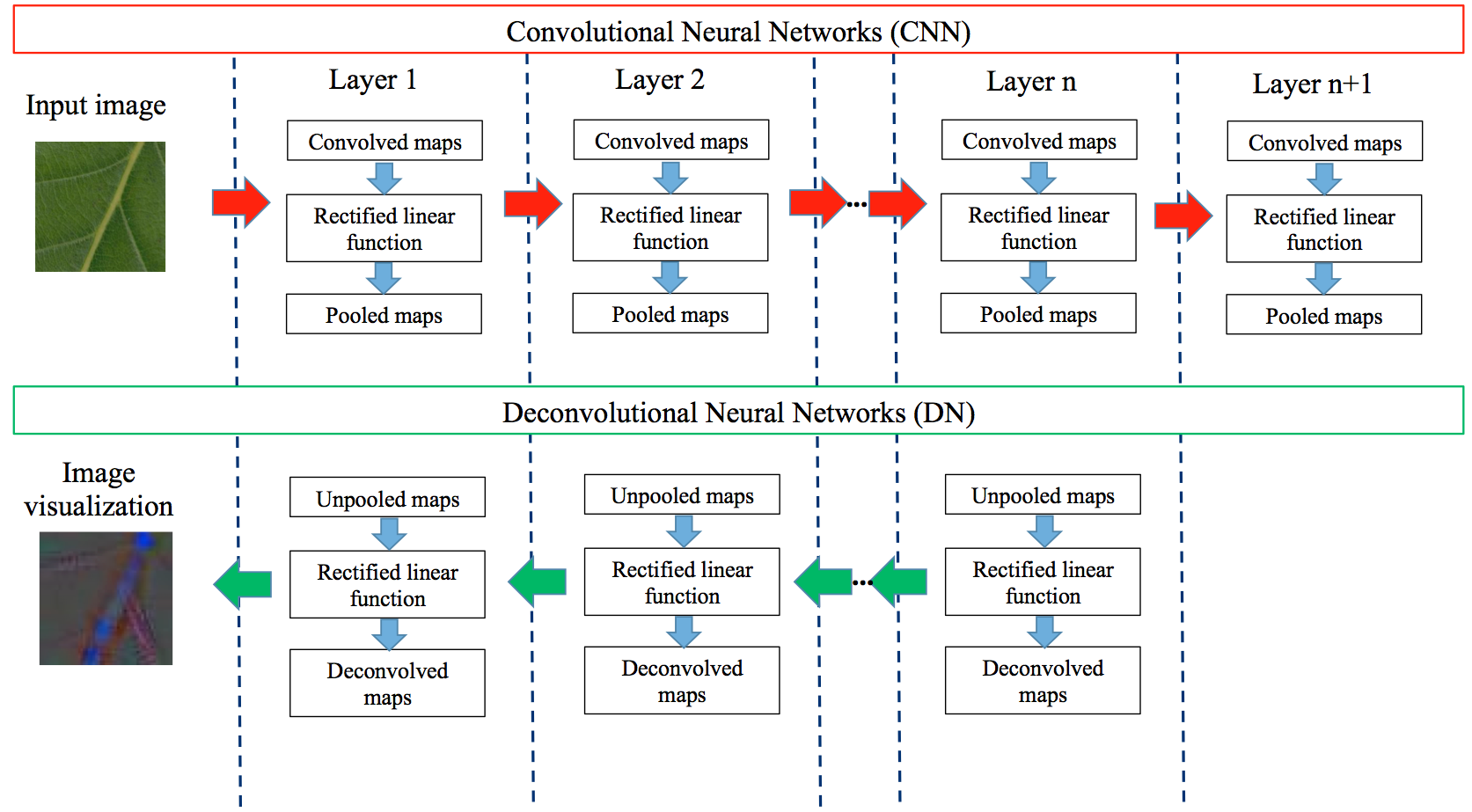}}
		\caption{Our deep learning framework in a bottom-up and top-down manner to study and understand plant identification.}
		\label{fig:deconv1}
	\end{figure}

\subsection{Convolutional Neural Network}
The CNN model used in this paper is based on the model proposed in \cite{krizhevsky2012imagenet} with ILSVRC2012 dataset used for pre-training. Rather than training a new CNN architecture, we re-used the pre-trained network due to a) recent work \cite{donahue2013decaf} reported that features extracted from the activation of a CNN trained in a fully supervised manner on large-scale object recognition works can be re-purposed to a novel generic task; 2) our training set is not large as the ILSVRC2012 dataset. Indicated in \cite{dong2014learning}, the performance of the CNN model is highly depending on the quantity and the level of diversity of training set, and finally c) training a deep model requires skill and experience. Also, it is time-consuming. 

For our CNN model, we perform fine-tuning using a 44 classes leaf dataset collected at the Royal Botanic Gardens, Kew, England. Thus, the final fully connected layer is set to have 44 neurons replacing the original 1000 neurons. The full model of our CNN architecture is depicted in Fig. \ref{fig:imagenet}. The first convolutional layer filters the  227$\times$227$\times$3 input leaf images with 96 kernels of size 11$\times$11$\times$3 with stride of 4 pixels. Then, the second convolutional layer takes the pooled feature maps from the first layer and convolved with 256 filters of size 5$\times$5$\times$48. Following this, the output is fed to the third and later to the fourth convolutional layer. The third and fourth convolutional layers  which have 384 kernels of size 3$\times$3$\times$256 and 384 kernels of size 3$\times$3$\times$192 respectively perform only convolution without pooling. The fifth convolutional layer has 256 kernels of size 3$\times$3$\times$192. After performing convolution and pooling in the fifth layer, the output is fed into fully-connected layers which have 4096 neurons. For the parameter setting, the learning rate multiplier of the filters and biases are set to 10 and 20, respectively.

\subsection{Deconvolutional Network}
\label{sec:decov}

The CNN model learns and optimises the filters in each layer through the back propagation mechanism. These learned filters extract important features that uniquely represent the input leaf image. Therefore, in order to understand why and how the CNN model operates (instead of treating it as a "black box"), filter visualisation is required to observe the transformation of the features, as well as to understand the internal operation and the characteristic of the CNN model. Moreover, we can identify the unique features on the leaf images that are deemed important to characterize a plant from this process. \cite{zeiler2011adaptive, zeiler2014visualizing} introduced multi-layered DN that enable us to observe the transformation of the features by projecting the feature maps back to the input pixel space. Specifically, the feature maps from layer $n$ are alternately deconvolved and unpooled continuously down to input pixel space. That is, given the feature maps, $Y_i^{(l-1)}$ as:
\begin{equation}
Y_i^{(l-1)} =\sum\limits_{j=1}^{m_1^{(l)}} {(K_{j,i}^{(i)})}^T \ast Y_j^{(l)}
\end{equation}

\noindent where layer $l$ be a deconvolutional layer and $K$ are the filters. 

	\begin{figure}[t]
		\centering
		\centerline{\includegraphics[height=0.65\linewidth, width=0.95\linewidth]{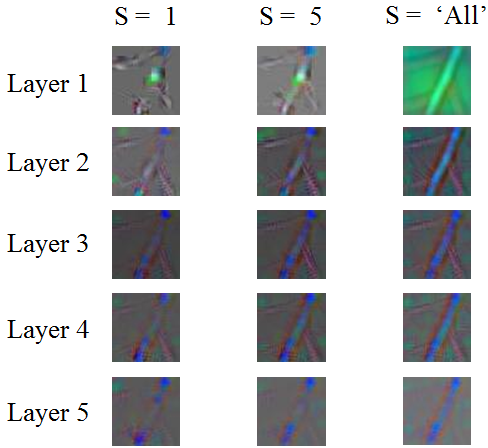}}
		\caption{Visualization (V1) strategy to understand how and why our CNN works/fails. Best viewed in colour.}
		\label{fig:deconv}
	\end{figure}
	
To visualize our CNN model, we employ a strategy named as {\bf V1} based on the DN approach \cite{zeiler2011adaptive, zeiler2014visualizing}. The purpose of V1 is to examine the overall highest activation parts across all feature maps for that layer $l$. So that, through the reconstructed image, we could observe the highly activated regions of the leaf in that layer. In order to do that, for all the absolute activations in that layer $n$, we consider only the first $S$ largest pixel value with the rest are set to zero and projected down to pixel space to reconstruct an image as defined:
 \begin{equation}
 Y_s^{(l-1)} =\sum\limits_{j=1}^{m_1^{(l)}} {(K_{j,i}^{(i)})}^T \ast Y_j^{(l)}  
 \end{equation} 
 
\noindent where $S = {1,2,.....,size(Y_j^{(l)})}$.  With this, we could observe the highly activated regions of the leaf in that layer. The visual results of S = 1, S = 5 and S = 'All' are illustrated in Fig. \ref{fig:deconv}. 

\section{Experimental Results}
\label{sec:illust}

\subsection{Data Preparation}
\label{sec:data}
A new leaf dataset, named as MalayaKew (MK) Leaf Dataset which consists of 44 classes, collected at the Royal Botanic Gardens, Kew, England are employed in the experiment. Samples of the leaf dataset is illustrated in Fig. \ref{fig:sample}, and we could see that this dataset is very challenging as leaves from different classes have very similar appearance. A data (D1) is prepared to compare the performance of the trained CNN. That is, we use leaf images as a whole where in each leaf image, foreground pixels are extracted using the HSV colour space information. To enlarge the D1 dataset, we rotate the each leaf images in 7 different orientations, e.g. $45^{\circ}$, $90^{\circ}$, $135^{\circ}$, $180^{\circ}$, $225^{\circ}$, $270^{\circ}$  and $315^{\circ}$. We then randomly select 528 leaf images for testing and 2288 images for training.

\begin{figure}[!t]
	\centering
	\includegraphics[height=0.55\linewidth, width=0.85\linewidth]{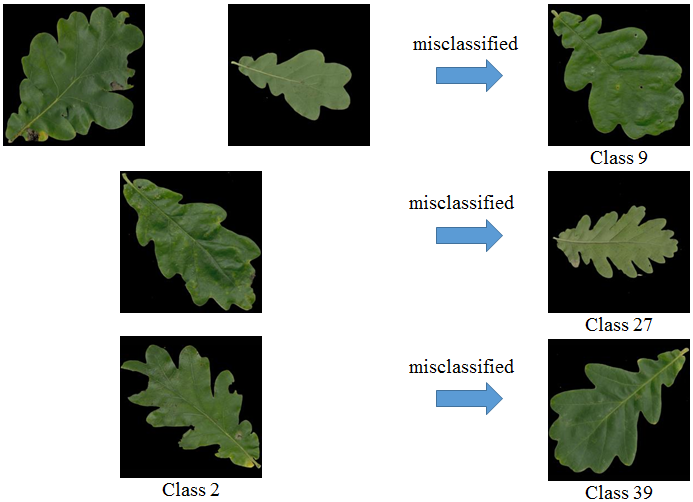}
	\caption{Failure analysis on our proposed CNN model in D1.}
	\label{fig:missclassifyD1a}
\end{figure}

\begin{figure}[!t]
	\centering
	\includegraphics[height=0.16\linewidth, width=0.89\linewidth]{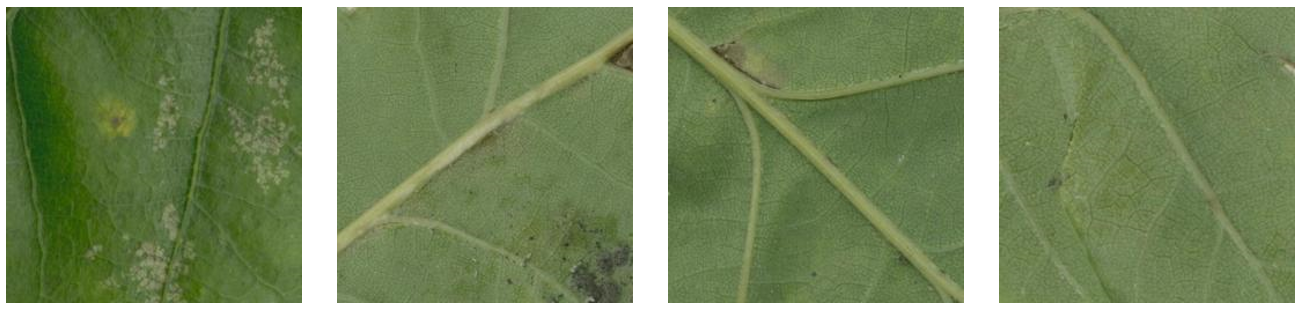}
	\caption{Failure analysis on our proposed CNN model in D2.}
	\label{fig:missclassifyD2}
\end{figure}

\begin{figure*}[!t]
	\centering
	\includegraphics[height=0.3\linewidth, width=0.9\linewidth]{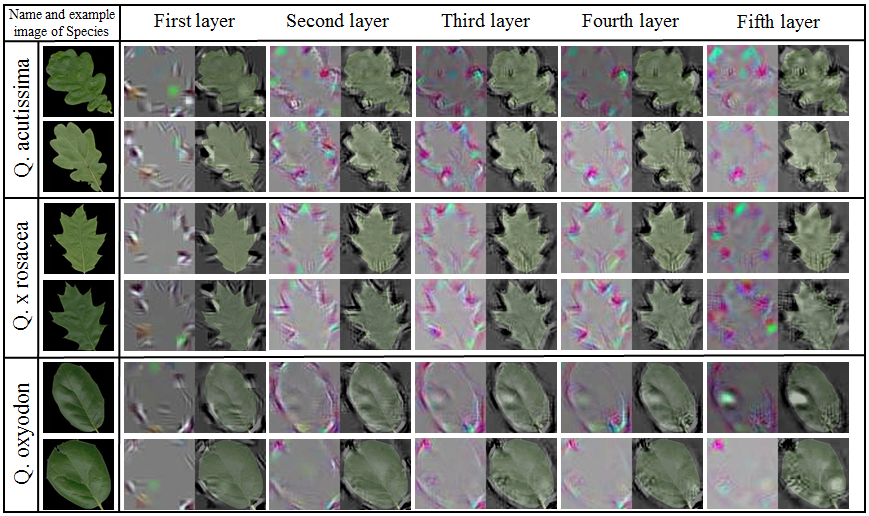}
	\caption{Feature visualisation using DN. It shows that shape (feature) is chosen in D1. Best viewed in colour.}
	\label{fig:qualitativeNewa}
\end{figure*}

\begin{figure*}[!t]
	\centering
	\includegraphics[height=0.3\linewidth, width=0.9\linewidth]{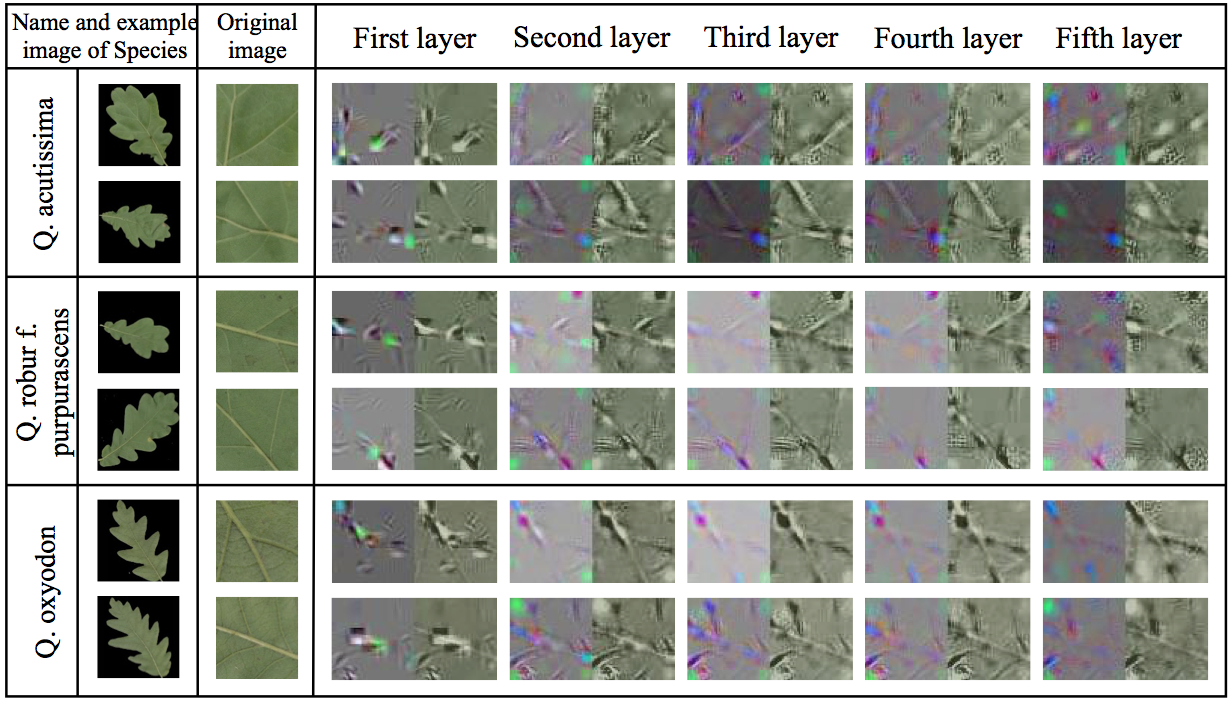}
	\caption{Feature visualisation using DN. It shows that venation and the departure between different order venations (feature) are chosen in D2. Best viewed in colour.}
	\label{fig:qualitativeNewb}
\end{figure*}

\subsection{Results and Failure Analysis - D1}

In this section, we present a comparative performance evaluation of the CNN model on plant identification. From Table \ref{table:result}, it is noticeable that using the features learnt from the CNN model (98.1\%) outperforms state-of-the-art solutions \cite{kumar2012leafsnap,yang2009linear,hall2015evaluation} that employed carefully chosen hand-crafted features even when different classifiers are used. We performed failure analysis and observed that most of the misclassified leaves are from Class 2(4 misclassified), follow by Class 23(3), Class 9 \& 27(2 each), and Class 38(1).  From our investigation as illustrated in Fig. \ref{fig:missclassifyD1a}, the {\it Q. robur f. purpurascens} (i.e Class 2) were misclassified as {\it Q. acutissima} (i.e Class 9) , {\it Q. rubra ‘Aurea’} (i.e. Class 27)  and {\it Q. macranthera} (Class 39), respectively; have almost the same outline shape as to Class 2. The rest of the misclassified testing images are also found to be misled by the same reason.

In order to further understand how and why the CNN fails, here we delve into the internal operation and behaviour of the CNN model via V1 strategy. We evaluate the one largest pixel value across the feature maps. Our observation from the reconstructed images in Fig \ref{fig:qualitativeNewa} shows that the highly activated parts fall at the shape of the leaves. So, we deduce that leaf shape is not a good choice to identify plants.

\begin{table}[ht]\footnotesize
	\caption{Performance Comparison on the MK Leaf Dataset with Different Classifiers. Note that, MLP = Multilayer Perceptron, SVM = Support Vector Machine, and RBF = Radial Basis Function.}
	\begin{tabular}{ | c || c | c | cp{cm} |}
		\hline
		Feature & Classifier  & Accuracy (\%) \\ \hline \hline
		From Deep CNN (D1)  & MLP  & 0.977 \\ \hline
		From Deep CNN (D1)  & SVM (linear)  & 0.981 \\ \hline
		From Deep CNN (D2) & MLP  & {\bf 0.995} \\ \hline
		From Deep CNN (D2)  & SVM (linear)  & 0.993 \\ \hline
		LeafSnap \cite{kumar2012leafsnap} & SVM (RBF) & 0.420\\ \hline
		LeafSnap \cite{kumar2012leafsnap} & NN & 0.589\\ \hline	
		HCF  \cite{hall2015evaluation} & SVM (RBF) & 0.716\\ \hline
		HCF-ScaleRobust \cite{hall2015evaluation} & SVM (RBF) & 0.665\\ \hline
		Combine \cite{hall2015evaluation} & Sum rule (SVM (linear)) & 0.951\\ \hline
		SIFT \cite{yang2009linear} & SVM (linear) & 0.588 \\ 
		\hline
	\end{tabular}
	\label{table:result}
\end{table}

\subsection{Results and Failure Analysis - D2}

Here, we built a variant dataset (D2), where we manually crop each leaf image in the D1 into patches within the area of the leaf (so that no shape is included). This investigation is two-fold. On one hand, we intend to know what is the precision of the plant identification classifier when the leaf shape is excluded ? On the other hand, we would like to find out if plant identification could be just done by patch of the leaf.  Since the original images range from 3000 $\times$ 3000 to 500 $\times$ 500, three different leave patch sizes (\ie~500 $\times$ 500, 400 $\times$ 400 and 256 $\times$ 256) are chosen. Similarly, we increase the diversity of the leaf patches by rotating them it in the same manner as to D1. We randomly select 8800 leaf patches for testing and 34672 leaf patches for training.

In Table \ref{table:result}, we can see that the classification accuracy of the CNN model, trained using D2 (99.6\%), is higher than using D1 (97.7\%). Again, we perform the visualisation via V1 strategy as depicted in Fig. \ref{fig:qualitativeNewb} to understand why the CNN trained with D2 has a better performance. From layer to layer, we notice that the activation part falls on not only the primary venation but also on the secondary venation and the departure between different order venations. Therefore, we could deduce that venation of different orders are more robust features for plant identification. This also agrees with some studies \cite{roth2001evolution,candela1999venation} highlighting that quantitative leaf venation data have the potential to revolutionize the plant identification task. Existing work that had employed venation to perform plant classification are \cite{runions2005modeling,clarke2006venation,cope2010extraction,larese2014multiscale,mullen2008artificial}. However, as opposed to these solutions, we automatically learn the venation of different orders, while they use a set of heuristic rules that are hard to replicate.

We also analysed the drawbacks of our CNN model with D2 and observe that most of the misclassified patches are from Class 9(18 misclassified), follow by Class 2(13), Class 30(5), Class 28(3) and Class 1 , 31 \& 42(1 each). The contributing factor of the misclassification seems to be the condition of the leaves, where the samples are noticeable affected by environmental factors such as wrinkled surface and insect damages. Example of such conditions are shown in Fig. \ref{fig:missclassifyD2}.

\section{Conclusion}
\label{conc}
This paper studied a deep learning approach to learn discriminative features from leaf images with classifiers for plant identification. From the experimental results, we justified that learning the features through CNN can provide better feature representation for leaf images compared to hand-crafted features. Moreover, we demonstrated that venation structure is an important feature to identify different plant species with performance of 99.6\%, outperforming conventional solutions. This is verified by analysing the internal operation and behaviour of the network through DN visualisation technique. In future work, we will extend the work to recognize in the wild.

\section*{Acknowledgment}
\label{ack}
This research is supported by the High Impact MoE Grant UM.C/625/1/HIR/MoE/FCSIT/08, H-22001-00-B00008 from the Ministry of Education Malaysia.

\bibliographystyle{IEEEbib}
\bibliography{refs}

\end{document}